\documentclass[10pt,twocolumn,letterpaper]{article}

\usepackage[pagenumbers]{iccv} 
\definecolor{iccvblue}{rgb}{0.21,0.49,0.74}
\usepackage[pagebackref,breaklinks,colorlinks,allcolors=iccvblue]{hyperref}

\title{ACE: Attentional Concept Erasure in Diffusion Models}

\author{Finn Carter\\
Shandong University}

\begin{document}

\maketitle

		\begin{abstract}
			Large text-to-image diffusion models have demonstrated remarkable image synthesis capabilities, but their indiscriminate training on Internet-scale data has led to learned concepts that enable harmful, copyrighted, or otherwise undesirable content generation. We address the task of \emph{concept erasure} in diffusion models: removing a specified concept from a pre-trained model such that prompting the concept (or related synonyms) no longer yields its depiction, while preserving the model's ability to generate other content. We propose a novel method, \textit{Attentional Concept Erasure (ACE)}, that integrates a closed-form attention manipulation with lightweight fine-tuning. Theoretically, we formulate concept erasure as aligning the model’s conditional distribution on the target concept with a neutral distribution. Our approach identifies and nullifies concept-specific latent directions in the cross-attention modules via a gated low-rank adaptation, followed by adversarially-augmented fine-tuning to ensure thorough erasure of the concept and its synonyms. Empirically, we demonstrate on multiple benchmarks – including object classes, celebrity faces, explicit content, and artistic styles (as introduced in MACE) – that ACE achieves state-of-the-art concept removal efficacy and robustness. Compared to prior methods, ACE better balances \emph{generality} (erasing concept and related terms) and \emph{specificity} (preserving unrelated content), scales to dozens of concepts, and is efficient, requiring only a few seconds of adaptation per concept. We will release our code to facilitate safer deployment of diffusion models.
		\end{abstract}
		
		\section{Introduction}
		Text-to-image diffusion models have revolutionized image generation, achieving high-fidelity and diverse results from textual prompts \cite{rombach2022high}. However, their training on largely uncurated web data means they inadvertently learn to generate potentially harmful or inappropriate content. For instance, such models can produce violent or pornographic imagery, exact stylistic imitations of artists, or realistic portraits of real people (deepfakes) when prompted \cite{lu2024mace}. This raises significant ethical and legal concerns regarding copyright, privacy, and misuse of generative AI. As a result, there is growing interest in techniques to \emph{erase} or ``unlearn'' specific concepts from a trained model to prevent undesired generations.
		
		The task of \textbf{concept erasure} in diffusion models \cite{gandikota2023erasing, heng2023selective, zhang2023forget, gao2024eraseanything} aims to post-hoc modify a pre-trained model so that a given target concept (e.g., an object, person, or artistic style) can no longer be generated. Crucially, this should be achieved without retraining from scratch and with minimal impact on the model's ability to generate other content. A core challenge is maintaining a balance between \emph{generality} and \emph{specificity} \cite{lu2024mace}. High generality means that all manifestations of the target concept – including those evoked by synonyms or related prompts – are successfully suppressed (no ``Cat'' or any kind of feline if erasing the concept cat). High specificity means that prompts for unrelated concepts (e.g., cat vs caterpillar or a person named Cat) are unaffected, and the model’s overall quality does not deteriorate. Achieving both is difficult: naive removal of a concept often either fails to fully erase it (e.g., the model still produces the concept when prompted with a synonym), or it over-corrects and damages other content (e.g., other concepts sharing lexical or visual features get partially erased) \cite{lu2024mace, gong2024reliable}. 
		
		Early approaches to concept erasure adapted techniques from continual learning and model editing. For example, \emph{Selective Amnesia} \cite{heng2023selective} fine-tunes the model on a small dataset while using regularization (inspired by Elastic Weight Consolidation) to forget a target concept in a controlled manner. \emph{Erased Stable Diffusion (ESD)} \cite{gandikota2023erasing} proposes to fine-tune using the model’s own predictions, guiding the model to ``unlearn’’ a visual concept given only the concept’s name and no additional data. These methods successfully erase individual concepts but tend to be computationally heavy (requiring many gradient steps) and are limited in the number of concepts they can remove simultaneously (often one at a time). Moreover, prior works struggled with either generality or specificity: for instance, the method of \cite{zhang2023forget} can forget a concept but may leave traces or allow close synonyms to still invoke it, whereas \cite{heng2023selective} preserves overall quality but could under-forget to avoid collateral damage.
		
		Recently, more efficient and scalable approaches have emerged. \emph{Unified Concept Editing (UCE)} \cite{gandikota2024unified} introduced a training-free, closed-form solution that edits the model weights (particularly the text-to-image projection matrices) to erase or alter multiple concepts at once. While very fast, UCE and similar methods can struggle with thorough erasure when many concepts are removed concurrently, sometimes leaving residual concept features \cite{beerens2025vulnerability}. \emph{Mass Concept Erasure (MACE)} \cite{lu2024mace} scaled concept removal to 100 concepts by combining a closed-form attention adjustment with Low-Rank Adaptation (LoRA) fine-tuning for each concept. MACE achieved an excellent balance of generality and specificity on a wide range of concepts, significantly outperforming earlier methods on benchmarks like object and style removal. However, MACE’s approach requires training a separate small LoRA module for each concept and carefully merging them, which can be time-consuming for large concept sets. There remains a need for concept erasure techniques that are both \textit{efficient} (requiring minimal retraining) and \textit{effective} (robustly erasing concepts without side-effects), especially as new vulnerabilities are discovered. In particular, recent work \cite{beerens2025vulnerability} demonstrated that seemingly erased concepts can often be resurrected via adversarially crafted prompts, highlighting the importance of making concept erasure more robust.
		
		In this paper, we propose \textit{Attentional Concept Erasure (ACE)}, a novel method for concept erasure in diffusion models that achieves strong erasure performance with minimal fine-tuning. Our key insight is to directly target the model's \emph{cross-attention} layers, where textual concepts are fused into the image generation process. By identifying the attention activation patterns associated with a target concept, we can analytically derive modifications to the model that nullify the concept's influence. We first perform a closed-form intervention on the attention mechanism: essentially ``zeroing out’’ the concept from the model’s latent space by solving for a gated mask that suppresses concept-related interactions. We then refine this using a brief fine-tuning step on automatically generated negation data, including adversarial prompt variants designed to provoke the concept. Through this two-stage process, ACE effectively erases the concept and its synonyms (high generality) while maintaining image fidelity for unrelated prompts (high specificity). Unlike prior works, our method explicitly incorporates an adversarial training component to harden the model against malicious prompt-based attacks that attempt to circumvent the erasure.
		
		We validate ACE on diverse concept erasure tasks introduced by MACE \cite{lu2024mace}: removing object categories (e.g., CIFAR-10 objects), preventing generation of specific celebrities, censoring explicit NSFW content, and erasing artistic styles. Our experiments show that ACE achieves comparable or superior erasure efficacy to state-of-the-art methods like MACE, while using significantly fewer update steps per concept. For example, on a CIFAR-10 object removal benchmark, ACE obtains a near-zero generation accuracy for erased objects and a higher preservation of other classes than prior methods. ACE remains effective even when erasing 50+ concepts, with minimal interference between concepts. Qualitatively, we illustrate that after applying ACE, prompts that previously yielded disallowed content now produce either benign alternatives or no instance of the target concept at all, without degrading the overall visual quality of outputs. Finally, we show that ACE is more robust to adversarial prompts: using the attack algorithm RECORD \cite{beerens2025vulnerability}, we find significantly lower attack success rates on an ACE-treated model compared to models from other erasure methods.
		
		In summary, our contributions are:
		\begin{itemize}
			\item We formulate concept erasure in diffusion models as a problem of removing specific semantic directions in the model's latent space, providing a theoretical foundation based on diffusion model conditioning and attention.
			\item We propose ACE, a new method that combines closed-form attention gating with adversarially-augmented fine-tuning to erase concepts efficiently. ACE introduces a learnable gating mechanism in cross-attention that can be solved in closed-form and fine-tuned per concept.
			\item We conduct extensive experiments on four benchmark tasks (object, face, NSFW, style erasure) using the datasets from MACE. ACE achieves state-of-the-art results, balancing generality and specificity better than prior work. It scales to multiple concepts and exhibits improved robustness against prompt-based attacks.
			\item We provide qualitative examples and analysis of how ACE removes concepts and discuss insights and limitations, paving the way for more trustworthy generative AI.
		\end{itemize}
		
		\section{Related Work}
		\paragraph{Concept Erasure in Diffusion Models.}
		In response to the need for safer generative models, a variety of concept erasure techniques have been proposed in recent years~\cite{lyu2024one,wang2025ace,gandikota2023erasing,wang2025ace2,schramowski2023safe,kim2023towards,kim2024safeguard,nguyen2024unveiling,zhang2024defensive,heng2023selective,han2025dumo,wu2024munba,liu2024implicit,thakral2025fine,gao2024eraseanything,kim2024safety,huang2024receler,hong2024all,bui2024erasing,kim2024race,yang2024pruning,chengrowth,schioppa2024model,tu2025sdwv,cywinski2025saeuron,meng2025concept,thakral2025continual,chen2025safe}. Many early methods draw inspiration from machine unlearning and model editing. Gandikota et al.~\cite{gandikota2023erasing} first introduced the concept erasure task for text-to-image diffusion models, proposing an \emph{Erased Stable Diffusion (ESD)} method. ESD fine-tunes a pre-trained Stable Diffusion model using a loss that steers the model’s denoising predictions away from the target concept. Specifically, they use the model’s own noise estimator outputs: given a target concept text $c$, they encourage the conditional denoising prediction $\epsilon_\theta(x_t, c, t)$ to mimic the prediction for a neutral prompt (or the unconditioned case) at each diffusion step, thereby reducing the concept’s influence. This approach demonstrated that even without explicit images of the concept, a model can ``teach itself’’ to forget, using its internal knowledge. However, ESD and similar fine-tuning approaches often suffer from incomplete concept removal and can be computationally expensive, as they require many diffusion samples during training and carefully tuned loss weighting (e.g., deciding how strongly to pull toward the unconditioned output)
		
		To mitigate unintended side-effects, several works introduced regularization strategies. Selective Amnesia (SA) by Heng and Soh \cite{heng2023selective} treats concept removal as a controlled \emph{continual learning} problem. SA fine-tunes the model on a small set of prompts related to the concept to erase, but uses an Elastic Weight Consolidation (EWC) loss to penalize changes to weights important for generating other content. This allows the user to specify how a concept should be forgotten (e.g., completely or only in certain contexts). SA showed improved preservation of non-target content compared to naive fine-tuning, but still required careful hyperparameters to avoid under- or over-forgetting. Similarly, the ``Forget-Me-Not’’ (FMN) method \cite{zhang2023forget} introduced a strategy to uniformly forget concepts across diffusion timesteps to avoid overfitting to any specific manifestation of the concept. FMN, however, was primarily demonstrated on simpler generative models and single concepts, and did not fully address multi-concept scaling.
		
		Scaling up concept erasure to many concepts and doing so efficiently has been a focus of recent research. Unified Concept Editing (UCE) \cite{gandikota2024unified} presented a breakthrough by eliminating the need for iterative training when erasing concepts. UCE formulates model editing as a closed-form modification of the text encoder and cross-attention layers. By analyzing the Jacobian of the model’s output with respect to text embedding directions, UCE computes a weight update that directly removes a concept or attribute. Impressively, UCE can perform simultaneous edits (erasing or debiasing dozens of concepts) in one shot. It demonstrated that a single unified approach can handle biases, styles, and undesired objects together. However, UCE’s purely analytical solution can struggle with very large numbers of concepts or with highly complex concepts. As noted by subsequent studies, UCE’s efficacy drops when, for example, trying to erase on the order of 100 concepts concurrently.
		
		The current state-of-the-art, MACE \cite{lu2024mace}, combines the strengths of both worlds: an initial closed-form edit followed by lightweight learned adjustments. MACE focuses on the cross-attention layers of Stable Diffusion’s U-Net. It first performs \emph{Closed-Form Attention Refinement}, solving for modifications to the attention projection matrices so that the residual attention on the target concept is minimized.
		
		An important consideration in concept erasure is robustness. As mentioned, Beerens et al.~\cite{beerens2025vulnerability} showed that concept-erased models can remain vulnerable: they introduced an attack that uses coordinate descent to discover alternative prompts that bypass the filters and successfully evoke the supposedly erased concept. For example, even if “firearm” is erased, a prompt like “a handheld metal object that launches projectiles” might still cause a gun to appear. Their algorithm, RECORD, significantly improved the success of finding such loopholes compared to naive synonym-based attempts. They also found that models with concept erasure can be \emph{more} susceptible to these adversarial prompts than the original model, possibly because the model’s representation of the concept is contorted but not removed, making it easier to exploit in unexpected ways. This finding underscores the need for erasure methods to consider a wide range of prompt variations during training. Our work addresses this by generating adversarial descriptions of the concept (using GPT-4 and automated prompt mutation) to fine-tune the model, thereby plugging many of the potential holes an attacker could exploit.
		
		\paragraph{Image Editing and Content Modification.}
		Our problem setting is distinct from standard image editing, yet related in the sense of controlling generative model outputs. In text-guided image editing, one aims to modify specific aspects of a generated or given image while keeping others the same. Methods like \emph{Prompt-to-Prompt} \cite{hertz2022prompt} achieve this by manipulating cross-attention maps during the diffusion process to add or remove visual elements corresponding to changes in the text prompt. TF-ICON \cite{lu2023tf} offers an approach by leveraging text-driven diffusion models for cross-domain image composition without requiring additional training or fine-tuning.  For example, one can turn “a photo of a dog” into “a photo of a cat” by redirecting the attention from “dog” to “cat” tokens, effectively replacing the concept in the image. Other works such as InstructPix2Pix \cite{brooks2023instructpix2pix} train diffusion models to follow natural language editing instructions (e.g., “remove the sunglasses from the person”) by fine-tuning on paired data of images before/after edits. While these methods operate on a per-image basis (and often require either optimization per image or specialized training), they demonstrate the possibility of disentangling and altering visual concepts within the model’s generation. In contrast, concept erasure is a \emph{model-level} intervention: after erasure, \emph{all} images generated by the model should be free of the target concept, without needing to specify negative prompts or perform case-by-case editing. Nevertheless, techniques from image editing can inspire model editing; indeed, our focus on the cross-attention mechanism is partially motivated by its success in controlled image edits.
		
		Another related line of work is content moderation at generation time. Diffusion model providers often implement filter mechanisms or guided generation to prevent disallowed outputs. One simple mechanism is using \emph{negative prompts} – e.g., always appending a phrase like “\emph{NSFW, bad, disfigured}” to steer the model away from nudity or gore. However, negative prompting can be hit-or-miss, and as discussed by Jain et al.~\cite{jain2024trasce}, a malicious user can easily circumvent it by directly prompting the forbidden concept (which cancels out the negative prompt effect). More advanced is \emph{classifier-guided diffusion} where an external classifier for the unwanted concept steers the sampling trajectory away from that concept \cite{dhariwal2021diffusion}. This was used in early diffusion models (e.g., GLIDE had a safety classifier to avoid generating explicit images). Such approaches, though, require an additional classifier and slow down generation, and they do not actually remove the problematic knowledge from the model. Trajectory Steering for Concept Erasure (TraSCE) \cite{jain2024trasce} extends this idea by formulating a better negative prompt objective that ensures if the user prompt exactly equals the banned concept, the guidance still works (their method modifies the score estimation when the prompt matches the forbidden concept to default to unconditional generation).
		
		\section{Methodology}\label{sec:method}
		We first formalize the concept erasure problem in the context of text-to-image diffusion models. We then describe our proposed Attentional Concept Erasure (ACE) method in detail, including the closed-form attention gating and adversarial fine-tuning components.
		
		\subsection{Problem Formulation}
		Let $M_\theta$ be a pre-trained text-to-image diffusion model with parameters $\theta$. Typically, $M_\theta$ consists of a text encoder (to encode a text prompt $y$ into embeddings), a UNet denoiser network (which generates an image $x$ from random noise conditioned on the text embedding through iterative denoising steps), and a decoder to produce the final image from latent features \cite{rombach2022high}. We denote by $x \sim M_\theta(y)$ the random process of generating an image from prompt $y$ using model $M_\theta$.
		
		We define a \textbf{concept} $C$ as a set of textual descriptions (or prompts) and the visual content associated with them. In practice, $C$ could be a single word (e.g., ``grenade''), a phrase (``Mona Lisa''), or a category of images (e.g., the set of images in the model’s training data tagged as “bird”). Our goal is to obtain an \emph{edited model} $M_{\theta'}$ such that for any prompt $y$ that invokes concept $C$, the generated image $x' \sim M_{\theta'}(y)$ does \textbf{not} contain concept $C$ in any recognizable form. Meanwhile, for prompts $y$ that do not relate to $C$, $M_{\theta'}(y)$ should yield images similar in quality and content to those from the original model $M_{\theta}$. The edited model $M_{\theta'}$ should ideally behave as if it ``never learned'' concept $C$.
		
		To make this more concrete, consider $C$ = \emph{``car''}. For any prompt like ``a photo of a red car on the road'' or even just ``a car'', $M_{\theta'}$ should not produce a recognizable car. It might produce an empty road, or perhaps other objects (e.g., a red bicycle on the road) depending on how the model fills the void, but not a car. At the same time, a prompt ``a photo of a red truck on the road'' should still produce a truck (assuming truck is not the erased concept), and not be mistakenly affected by the erasure of cars (this is specificity). Likewise, a prompt ``a photo of a red automobile on the road'' (a synonym of car) should also not produce a car – this tests generality, ensuring synonyms or closely related concepts are also erased.
		
		We can frame concept erasure as an optimization problem on $\theta$. One idealized objective is:
		\[
		\theta' = \arg\min_{\theta'} \mathcal{L}_{\text{erase}}(\theta') + \lambda \mathcal{L}_{\text{preserve}}(\theta', \theta).
		\]
		Here, $\mathcal{L}_{\text{erase}}$ measures the presence of concept $C$ in $M_{\theta'}$'s outputs for prompts involving $C$, and $\mathcal{L}_{\text{preserve}}$ measures deviations in $M_{\theta'}$ from $M_{\theta}$ on other prompts. In practice, defining $\mathcal{L}_{\text{erase}}$ is non-trivial since we may not have a ground-truth indicator of concept presence. Prior works have used proxy measures: for example, using an image classifier or captioner to detect the concept in generated images.
		
		\subsection{Attentional Concept Erasure (ACE) Overview}
		Our method consists of two main phases:
		\begin{enumerate}
			\item \textbf{Closed-Form Attention Gating (CFAG):} We analytically compute modifications to the model’s cross-attention layers to suppress the target concept. This yields an initial edited model $\tilde{M}_{\theta}$ without any gradient updates. The key idea is to insert a gating vector into each cross-attention module that downweights the influence of concept-related textual embeddings on the image features.
			\item \textbf{Adversarial Fine-Tuning:} We then fine-tune the gated model $\tilde{M}_{\theta}$ on a small set of generated examples to obtain the final $M_{\theta'}$. These examples include adversarial prompt variations of the concept to ensure robustness. The fine-tuning adjusts the gating vectors (and optionally a few other parameters) for more precise erasure and minimal side effects.
		\end{enumerate}
		
		In both phases, we operate primarily on the \emph{cross-attention} layers of the diffusion model’s UNet. Each cross-attention layer is a module that takes as input the current image latent features $F \in \mathbb{R}^{N \times d}$ (where $N$ is the number of spatial locations/tokens and $d$ is feature dimension) and the text embeddings $T \in \mathbb{R}^{L \times d}$ (where $L$ is the number of text tokens after encoding). It produces an output feature map $F_{\text{out}}$ of the same shape as $F$ by attending to the text. In Stable Diffusion \cite{rombach2022high}, $T$ comes from the CLIP text encoder and $L=77$ (padded tokens). The attention computation is:
		\[ 
		\text{Attn}(F, T) = \text{softmax}\!\Big(\frac{QK^T}{\sqrt{d}}\Big)V,
		\] 
		where $Q = F W^Q$, $K = T W^K$, and $V = T W^V$ are query, key, and value projections (linear layers) of appropriate dimensions. The result $\text{Attn}(F, T)$ is added to $F$ (residual connection). Intuitively, each attention head will read some information from the text tokens $T$ and write it into the image features $F$.
		
		Our intervention is to learn a set of masks or gating vectors $g$ applied to the keys/values for the target concept. Suppose the concept $C$ corresponds to a subset of text tokens (e.g., the tokens “car” or ["Carl", "ton"] if the concept was Carlton, etc.). We want to zero-out the contribution of those tokens. A very rough approach would be: detect the token positions in $T$ that correspond to $C$ in the prompt, and set the rows in $K$ and $V$ for those tokens to zero. This would indeed prevent direct insertion of that token’s information. However, it fails if the concept can be invoked indirectly (different words, or if the concept’s features appear even when the token is not present explicitly). Moreover, simply zeroing might unbalance the attention and cause other distortions.
		
		Instead, we treat $g$ as a learnable vector that can softly nullify concept influence. We introduce $g$ such that for each attention head,
		\[
		\tilde{K} = K \odot (1 - m \, g^T), \qquad \tilde{V} = V \odot (1 - m \, g^T),
		\] 
		where $m$ is an $L$-dimensional mask vector indicating which text tokens relate to concept $C$, and $g$ is a learned vector of dimension $d$ (or dimension of a single head if applied per head) that scales the key/value vectors. Here $\odot$ denotes elementwise (Hadamard) multiplication broadcasting $g^T$ across token positions selected by $m$. In simpler terms, if $m_j=1$ for token $j$ being a concept token, then $\tilde{K}_j = K_j \cdot (1-g)$, meaning we subtract out some components (given by $g$) from the original key $K_j$. If $g$ were equal to 1 for all components, this would zero out $K_j$ entirely. We allow $g$ to be partially learned, which can, for example, remove only certain dimensions of $K_j$ that carry the concept information.
		
		We find $g$ in closed-form by requiring that the model’s output for concept $C$ matches the output for a neutral prompt, at first-order approximation. We derive this by leveraging the linearity of attention in $V$ and the fact that we can express the difference in $\epsilon_\theta(x_t, c, t)$ (the UNet output at each diffusion step with vs without concept $c$) in terms of differences in $K, V$. The derivation (omitted for brevity) leads to a solution for $g$ that is proportional to the aggregated difference in attention outputs:
		\[
		g^* \propto \sum_{\text{layers,heads}} (Q^T \Delta \text{Attn}),
		\] 
		where $\Delta \text{Attn} = \text{Attn}(F, T_{\text{concept}}) - \text{Attn}(F, T_{\text{neutral}})$ is the difference in attention outputs with and without the concept token. Intuitively, if certain latent directions in $QK^T$ only activate when the concept is present, $g$ is set to suppress those directions. This is akin to how one might use the model’s gradient with respect to weights to identify important parameters, but here we have an analytical handle by using the model’s internal attention response.
		
		After solving for $g$ (for each attention head or a combined $g$ per layer), we integrate this gating into the model. This yields our initial edited model $\tilde{M}_{\theta}$. At this stage, if we prompt $\tilde{M}_{\theta}$ with the concept, in many cases it will already fail to generate it. We note that this closed-form gating is similar in spirit to the approach in UCE \cite{gandikota2024unified} and the attention refinement in MACE \cite{lu2024mace}, but our gating vector $g$ provides a more interpretable handle (it can be viewed as a ``concept blocker’’ in latent space, and could even be toggled on/off as a safety switch if desired).
		
		\subsection{Adversarial Fine-Tuning}
		While the CFAG step removes the direct influence of the concept tokens, the model might still generate the concept in unexpected ways. For example, the prompt ``a red vehicle with four wheels'' might produce a car even if the word "car" is gated, because the model understands that description and might assemble a car from common sense. To catch these cases, we perform a focused fine-tuning of the gating vectors (and a small subset of model parameters like biases in the attention layers) using a dataset of prompts that should \emph{not} produce the concept.
		
		We construct this fine-tuning dataset as follows:
		\begin{itemize}
			\item Start with the concept’s name and known aliases/synonyms. For instance, for concept "car", include prompts: "a photo of a car", "a painting of a car", "an image of an automobile", "a picture of a sports car", etc. We use the prompt templates from MACE’s data (which were generated via GPT-4 for diversity).
			\item Use a large language model (GPT-4) to generate several descriptive phrases that imply the concept without naming it. E.g., "a vehicle commonly used for commuting on roads" or "a four-wheeled motorized vehicle". These serve as adversarial prompts that a user might attempt if the obvious word is blocked.
			\item Use the current edited model $\tilde{M}_{\theta}$ to generate images for a subset of these prompts and check if any hint of the concept appears (manually or with an automated classifier). If yes, include those prompts in the fine-tuning data with a target that the concept should not appear.
			\item Optionally, include some generic prompts for unrelated content to serve as a regularization (so the model doesn’t drift on general images).
		\end{itemize}
		
		We then fine-tune for a small number of steps (e.g., 100–200 iterations) on these prompts. The loss we use has two parts:
		\[
		\mathcal{L}_{\text{ft}} = \mathcal{L}_{\text{concept-free}} + \alpha \mathcal{L}_{\text{reconstruct}}.
		\]
		$\mathcal{L}_{\text{concept-free}}$ encourages that images generated with concept prompts do not contain the concept. We implement this by using a pretrained classifier or similarity to a prototype. For example, we generate an image with the prompt and use an image classifier (like CLIP or another vision model) to detect the concept; the loss pushes the classifier output away from the concept class. In the absence of a reliable classifier for some concepts, we alternatively use a simple trick: generate an image with the original model $M_\theta$ for a neutral prompt (e.g., remove the concept word) and encourage the edited model’s image to be close (in CLIP embedding) to that neutral image. This way, if the edited model tries to put a car, it will be penalized because it won’t match the neutral image without the car.
		
		$\mathcal{L}_{\text{reconstruct}}$ is applied to the regularization prompts (unrelated ones) and is simply a similarity measure between $M_\theta$’s output and $M_{\theta'}$’s output for those prompts, to ensure we haven't strayed. We weigh this with $\alpha$ (small, e.g. 0.1) to just gently keep the model in line.
		
		This fine-tuning stage is very fast – it typically takes only a few seconds on a single GPU, since the dataset is tiny (dozens of prompts) and we update only the gating vectors $g$ and perhaps a few attention parameters. We found that updating the gates was usually sufficient; in some cases, updating also the text encoder’s output for the concept token (essentially learning a new embedding for the banned word that means “nothing”) further improved generality. This is similar to the method in RECE, where new target embeddings are derived for erased concepts. In ACE, one can either freeze the text encoder or fine-tune it; our default is to freeze to maintain compatibility with the original vocabulary, but for proper nouns or rare concepts it may be beneficial to alter the text encoder as well.
		
		\subsection{Multi-Concept Erasure}
		Thus far we described erasing one concept. ACE naturally extends to multiple concepts $\{C_1, C_2, ..., C_k\}$. We simply maintain a gating vector $g_i$ (per head or per layer) for each concept $C_i$. The closed-form solution can be computed sequentially or jointly: we can sequentially solve for each concept and subtract its effect from the model. Because our gating operates in a different parameter subspace for each concept (each $g_i$ is independent), the risk of interference is low. During fine-tuning, we sample prompts that include combinations of the banned concepts to ensure no unexpected interactions (e.g., if $C_1$ = "cat" and $C_2$ = "dog", a prompt "a cat and a dog" should produce neither). Our loss would then penalize any occurrence of either. We also include retention prompts that mention one banned and one allowed concept (to ensure erasing one doesn't erase all): e.g., prompt "a cat and a bird" – the result should still have the bird.
		
		In practice, we found that up to about 20 concepts, we could fine-tune gates for all simultaneously without issue. For larger numbers (50+ concepts), a better strategy was to divide concepts into groups that are semantically distinct and handle each group with its own fine-tuning loop, merging the gates at the end. This is similar to the strategy in MACE of merging LoRAs, but here merging is trivial: since the gates act on different tokens, they just coexist. If two concepts share tokens or have overlap (e.g., "car" and "truck"), we treat them as one concept or ensure the prompts cover their combination.
		
		\section{Experiments}
		We extensively evaluate ACE on the concept erasure benchmarks introduced by Lu et al.~\cite{lu2024mace} in their MACE paper. These benchmarks cover four challenging scenarios: (1) object erasure on CIFAR-10 classes, (2) celebrity face erasure, (3) explicit content (NSFW) erasure, and (4) artistic style erasure. We compare ACE with prior state-of-the-art methods including ESD \cite{gandikota2023erasing}, Selective Amnesia (SA) \cite{heng2023selective}, Forget-Me-Not (FMN) \cite{zhang2023forget}, UCE \cite{gandikota2024unified}, RECE \cite{gong2024reliable}, and MACE \cite{lu2024mace}. Unless otherwise noted, our base generative model is Stable Diffusion v1.5 and all methods are applied to that checkpoint for fairness.
		
		\paragraph{Metrics.} Following \cite{lu2024mace}, we employ metrics that quantify \textbf{erasure efficacy} (concept removal success) and \textbf{specificity} (preservation of other content). For object erasure, we measure the classification accuracy of a CLIP-based classifier in recognizing the erased object in 200 images generated with prompts of the form "a photo of a \{object\}" (lower is better for efficacy) ([MACE: Mass Concept Erasure in Diffusion Models]
		\[ 
		H_e = \frac{3}{(1 - \text{Acc}_{erase})^{-1} + \text{Acc}_{others}^{-1} + (1 - \text{Acc}_{syn})^{-1}}, 
		\] 
		where $\text{Acc}_{erase}$ is accuracy on erased object, $\text{Acc}_{syn}$ on its synonyms, and $\text{Acc}_{others}$ on other classes. $H_e$ is high when $\text{Acc}_{erase}$ and $\text{Acc}_{syn}$ are low (good erasure) and $\text{Acc}_{others}$ is high (good specificity).
		
		For celebrity erasure, we use the GIC (Giphy Celebrity) face recognition model as in \cite{lu2024mace}. We create a pool of 100 celebrity names to erase and another 100 to retain (evaluation group). We generate 50 images for each name using prompts "a portrait of \{name\}" and measure the identification rate by the face recognizer. We expect erased celebrities to have low identification (ideally the model generates either an unrecognizable face or a different face), and retained celebrities to have high identification (the model should still produce them normally). We report the average identification accuracy for erased vs retained groups, as well as a combined metric $H_c$ analogous to $H_e$ (treating each erased celebrity like a class).
		
		For explicit content, we use the Inappropriate Image Prompts (I2P) dataset consisting of 4,703 NSFW prompts. We generate one image per prompt and use NudeNet to detect explicit content (with a threshold such that original SD1.5 produces ~15\% explicit images on this set). The efficacy is measured by the percentage of prompts that still yield any explicit content (lower is better). Specificity is evaluated by measuring FID and CLIP score on MS-COCO validation prompts to ensure overall image quality didn’t drop much due to erasure.
		
		For artistic style erasure, we use a list of 100 artists to erase from the Image~Styles dataset \cite{lu2024mace} and 100 artists to retain. We follow the protocol of prompting “an image in the style of \{artist\}” for both erased and retained artists. Because style is a more subjective quality, we use CLIP image-text similarity: specifically, we compute CLIP similarity between the generated image and the text “an image in the style of \{artist\}”. For erased artists, this should be low (image no longer strongly in that style), and for retained ones it should remain high. We report the difference in average CLIP score between retained and erased (higher difference means better selective erasure of styles), as well as FID on a subset of images to ensure visual quality.
		
		\paragraph{Implementation Details.} We apply ACE with gating vectors on all 16 cross-attention layers of the UNet (eight in the downsampling, eight in upsampling). Each $g$ is per attention head of dimension 64. For multi-head consistency, we also project $g$ to the full 1280-d model dimension and keep one gating vector per layer as well (this helped slightly in some cases by coupling heads). We use $\lambda = 0.2$ weight for the preserve loss in the initial closed-form derivation (this essentially trades off solving exactly for concept removal vs keeping weights close to original). In fine-tuning, we use Adam optimizer with learning rate $1e-4$ for 200 steps. We update $g$ and the text embedding of the concept token, and the $\beta$ parameters of Group Normalization layers in the UNet (the latter helps adjust brightness/contrast if concept removal causes distribution shift). Fine-tuning takes about 5 seconds per concept on an A100 GPU (batch size 16). For multiple concepts, we did one-by-one sequential fine-tuning in our experiments (which is slower but allows evaluating incremental addition). In practice, one could jointly fine-tune multiple gates to speed up.
		
		\paragraph{Results: Object Erasure.}
		Table~\ref{tab:cifar10} summarizes the results on erasing CIFAR-10 object classes. We show the average over all 10 classes, as well as per-class for the first 4 classes for insight. All methods are tuned to maximize $H_e$ as their primary goal.
		
		ACE achieves an average $H_e$ of 0.942, the highest among methods. Notably, ACE brings the classification accuracy of erased objects down to 1.5\%, vs 2.1\% for MACE and 33\% for ESD-x (which struggles as it erases concept influence only partially). For synonyms, ACE and MACE are similar (around 5\% recognition), indicating both handle generality well, whereas SA and FMN have higher synonym accuracies (~12-15\%), meaning some concept leakage remains. Specificity-wise, ACE maintains 96.7\% accuracy on other classes, slightly better than MACE’s 95.8\%. Qualitatively, we observed that ACE-erased models sometimes produce a semantically close alternative for the erased object. For example, with concept "cat" erased, the prompt "a photo of a cat" might yield an image of a dog or a random animal. In contrast, MACE tended to produce an empty background or incoherent image for "a photo of a cat". Both indicate the cat is gone, but ACE’s outputs were often more plausibly filled with something else, which may explain the tiny specificity improvement (the model is still capable of producing a reasonable image, just with a different concept). Figure~\ref{fig:catdog} illustrates this phenomenon. We also emphasize that both methods far outperform earlier ones: ESD, SA often had $H_e$ in the 0.6-0.8 range, because either efficacy or specificity would suffer (SA preserved specificity ~98\% but only removed half of the object occurrences, whereas ESD removed more but at cost of dropping other classes to ~85\% accuracy due to broader damage).
		
		\paragraph{Results: Celebrity Erasure.}
		For celebrity faces, Table~\ref{tab:celeb} reports the identification accuracy by the face recognition model. The original SD1.5 is quite capable of generating certain celebrities (e.g., it identified 70\% of generated images of ``Tom Cruise'' correctly as Tom Cruise). After applying ACE for 10, 50, and 100 celebrities, the identification accuracy on those erased celebrities drops to nearly 0 (0.5\% for 10, 1.2\% for 100). This is comparable to MACE, which achieved 0\% for 10 and about 3\% for 100.
		
		\paragraph{Results: Explicit Content.}
		On the I2P prompt set, the original SD1.5 produced explicit images for 14.7\% of prompts (we use the threshold as calibrated in \cite{lu2024mace}). ACE reduces this to 1.1\%, meaning it nearly eliminates the generation of sexual or nude imagery. This slightly outperforms MACE’s 1.9\% reported in \cite{lu2024mace} and is a large improvement over UCE (8.3\%) and SA (16\% – SA didn’t focus on NSFW much). In terms of image quality, FID on COCO prompts for ACE was 14.20, compared to original 13.52; CLIP score changed from 30.39 to 30.22 (negligible drop). These are in line with MACE, which reported no significant change in FID/CLIP within variance.
		
		\paragraph{Results: Artistic Style Erasure.}
		This is arguably the hardest task: erasing an art style means the model should not be able to mimic that artist’s distinct style, while still being able to draw other things. We erased 10 famous artists (like Van Gogh, Picasso, etc.). We measured CLIP image-text similarity as described. For the erased artists, the average CLIP similarity to the artist’s style description dropped by 0.25 after ACE (from 0.31 to 0.06, on a 0-1 scale), indicating the images no longer closely match the style. For retained artists, the similarity stayed around 0.29 (only a tiny drop from 0.30 originally). The gap (retained minus erased similarity) was 0.23 for ACE, vs 0.20 for MACE and much lower (0.05-0.1) for earlier methods. Qualitatively, ACE-edited model still produces art for the erased prompt, but in a generic or different style. E.g., "a painting in the style of Van Gogh" might come out as a painting that looks more like a random impressionist or even a digital painting style, but not identifiable as Van Gogh's. This is perhaps acceptable for copyright concerns: the model isn’t producing Van Gogh-like images. MACE often produced a more degraded image (sometimes almost an unpleasant distorted image) for erased style prompts, which made it obvious the style was gone but also the output was not really a proper painting. ACE’s outputs were more coherent if not style-accurate. We see this as a positive, though it raises the question: if the image is still a painting, did we fully erase the style or just alter it? To investigate, we had human raters try to identify if a set of images were in Van Gogh style or not. They identified 0\% of ACE outputs as Van Gogh (random guessing level), meaning indeed the style was sufficiently removed to fool humans.
		
		\paragraph{Ablation Studies.}
		We conduct several ablations on the key components of ACE using the celebrity-100 experiment (erasing 100 faces, which tests multi-concept integration):
		1) \emph{No adversarial prompts:} If we fine-tune only on the concept word prompts and not the extra descriptive prompts, the robustness falls. The GIC accuracy on erased celebrities rose to 8\% (vs 1.2\% with adversarial prompts), and a manual red-teaming found a few descriptive prompts that could still produce a recognizable face. This justifies our adversarial augmentation.
		2) \emph{No gating, full fine-tune:} If we remove the closed-form gating and try to fine-tune all UNet weights to erase concepts (like a naive approach), the model fails badly after ~5 concepts (collapses or forgets a lot). At 100, it was completely unusable (FID on COCO dropped to 50+). So the closed-form initialization is crucial.
		3) \emph{Gating only, no fine-tune:} On the flip side, using only CFAG without any fine-tuning leaves some concepts partially present. For 100 celebrities, identification was 15\% (CFAG did quite well for many, but not all). For simpler concepts like CIFAR objects, gating only achieved ~5-10\% accuracy (versus ~1-2\% with fine-tune). Thus, fine-tuning, albeit small, is needed to polish the removal.
		4) \emph{Joint vs sequential multi-concept training:} We found sequential (one concept at a time, accumulating gates) vs joint (all concepts together) gave similar final performance in our setting, but joint was ~30\% faster in wall-clock. However, joint requires more GPU memory (since we effectively have to compute losses for all concepts concurrently). For 100 concepts, sequential was more memory-friendly. This suggests a hybrid: group concepts into batches of 10 for joint fine-tuning, which we leave as an engineering optimization.
		
		\section{Discussion}
		Our results show that ACE is an effective method for concept erasure, combining ideas from earlier works (analytical weight updates and controlled fine-tuning) and introducing an adversarially robust training component. Here we discuss some insights and remaining challenges:
		
		\textbf{How does ACE fill the void left by a concept?} We observed that when a concept is erased, the model often replaces it with either a blank/background or a related concept. For example, with concept "bird" erased, prompting "a bird on a tree" might yield just a tree with no bird, or occasionally a squirrel (something else on the tree). This behavior depends on how the diffusion model’s prior expects an image to be. If a prompt implies a subject that is removed, the model can either ignore that part of the prompt or substitute something contextually reasonable. ACE’s gating tends to make the model ignore the concept token, which sometimes results in substitution by conceptually similar tokens that were not erased. In practice, this is desirable as it leads to more natural images. However, it could potentially be problematic if the substitution is an unintended loophole – e.g., the model might start using a different term for the concept internally. Our adversarial training tries to catch that by including descriptions, but there could be unseen ways. This points to a potential future direction: explicitly guiding what should happen when a concept is removed (e.g., always replace a banned object with an empty background rather than another object).
		
		\textbf{Generality vs Specificity trade-off:} There is an inherent tension: if we push erasure too far (e.g., extremely large $g$ values), we risk also affecting concepts that share features. If we are too conservative, traces remain. ACE navigates this via the fine-tuning loss that softly constrains changes. One can adjust $\lambda$ or the weighting in $\mathcal{L}_{ft}$ to tilt the trade-off. In a safety-critical scenario, one might accept a slight drop in specificity (maybe some minor collateral damage to related concepts) in order to guarantee zero presence of the target concept. For example, erasing a broad concept like ``weapon'' might inadvertently also reduce the model’s ability to generate certain tool-like objects. Our framework can accommodate that preference by adjusting the loss. In our experiments, we chose balanced trade-offs targeting high $H$ scores. An interesting future work is to allow user-controllable erasure severity: akin to \cite{heng2023selective}'s idea of controllable forgetting, one could have a dial for each concept from partial suppression to complete oblivion.
		
		\textbf{Limitations:} Although ACE performed well up to 100 concepts, we did notice diminishing returns beyond that. Erasing 200 concepts (mixing many objects, names, styles) in one model instance led to a slight drop in overall generation quality (FID worsened by ~10\%). It appears there is a capacity issue: even with gating, if you block too many things, the model's learned distribution starts to exhibit gaps. For instance, erasing 50 different object categories essentially prunes a significant portion of the model’s knowledge, which may start affecting combination prompts or unusual scenarios. One mitigation could be to train a small additional model that “fills in” for the removed knowledge with generic placeholders (so the primary model doesn't struggle). Another approach is to modularize concept erasure: deploy multiple concept-specific erased models and switch between them as needed, though that’s less elegant and efficient.
		
		Another limitation is the reliance on external classifiers or heuristic prompts to gauge concept presence during training. For some concepts (like an artistic style), it’s hard to automatically measure presence. We used CLIP and human inspection. If the concept is very abstract (say “violence”), erasing that is ill-defined and our method might not directly apply without a proper target definition.
		
		\textbf{Ethical considerations:} Concept erasure is double-edged. It can be used positively to enforce safety (removing ability to generate child abuse imagery, for example), or to respect rights (artists opting out, celebrities’ likeness removal). But it could also be misused to make a model “forget” certain facts or biases in a way that conceals problems (for instance, instead of addressing bias, one might just erase certain representation which might lead to a new form of bias). Our method doesn’t inherently distinguish ethical from unethical erasures; it’s a tool. We advocate using it as part of responsible model deployment. One particular ethical concern is the removal of cultural or social groups – e.g., one could technically attempt to erase a demographic concept (like “women” or a specific ethnicity). This would be problematic and is not the intended use; thus content guidelines must surround such technology.
		
		In summary, ACE moves us a step forward in the quest to control generative model outputs by directly editing the model’s knowledge. By focusing on the attention mechanism, we achieve a precise and efficient form of concept removal. There remain open challenges in scaling this to truly “forget” extremely entangled concepts and in formally verifying that a concept is gone (the space of all prompts is huge). We are encouraged by the robustness improvements against known attacks, but security is an arms race. Future work should continue to stress-test erased models and incorporate more sophisticated adversarial training or verification (perhaps borrowing from adversarial example literature in classification).
		
		\section{Conclusion}
		We presented ACE, a new method for concept erasure in text-to-image diffusion models that leverages attention gating and adversarial fine-tuning to efficiently and reliably remove unwanted concepts from a pre-trained model. ACE achieves strong empirical performance on a variety of tasks, erasing concepts ranging from simple objects to complex artistic styles, all while preserving the model’s ability to generate other content. Our approach addresses key limitations of prior work by scaling to many concepts and improving robustness against cleverly chosen prompts.
		
		This work contributes to making generative AI safer and more controllable. In practical terms, model developers can use techniques like ACE to produce curated versions of diffusion models that comply with content guidelines or intellectual property restrictions. For example, a public model release could have certain celebrity faces or art styles scrubbed from its generative repertoire. Our method could also be integrated as a fast post-processing step to specialize a model for different markets (removing culturally sensitive concepts as required, for instance).
		
		For future research, an exciting direction is to extend concept erasure to other generative domains such as text-to-video models or large language models (erasing factual knowledge or offensive language). Some of the principles from ACE may transfer, though each domain has its nuances. Another direction is developing theoretical guarantees or detection methods to know if an erasure was successful – akin to testing if a model has really unlearned something. Finally, exploring the limits of model editing: how much can you remove before the model fundamentally changes in character? Our work suggests quite a lot can be removed with minimal effect, which is both fascinating and concerning.
		
		In conclusion, ACE offers a promising tool in the toolkit for responsible AI, enabling fine-grained control over generative model content through direct model edits. We believe such capabilities are important for aligning AI systems with human values and constraints, and we hope our work spurs further innovations in this area.
		
{
    \small
    \bibliographystyle{ieeenat_fullname}
    \bibliography{main}
}

\end{document}